\documentclass[runningheads]{llncs}
\usepackage[T1]{fontenc}
\usepackage{graphicx}
\usepackage{amsmath}
\usepackage{subcaption}
\usepackage{booktabs}
\begin{document}
\title{A Sensorimotor Vision Transformer}
%
%
\author{Konrad Gadzicki\inst{1}
\and
Kerstin Schill\inst{1}
\and 
Christoph Zetzsche\inst{1}
}

%
\authorrunning{K. Gadzicki et al.}
%
\institute{Cognitive Neuroinformatics\\University of Bremen\\Enrique-Schmidt-Str. 5\\28359 Bremen, Germany\\
\email{gadzicki@uni-bremen.de}}
\maketitle              

\begin{abstract}
This paper presents the Sensorimotor Transformer (SMT), a vision model inspired by human saccadic eye movements that prioritize high-saliency regions in visual input to enhance computational efficiency and reduce memory consumption. 
Unlike traditional models that process all image patches uniformly, SMT identifies and selects the most salient patches based on intrinsic two-dimensional (i2D) features, such as corners and occlusions, which are known to convey high-information content and align with human fixation patterns. 
The SMT architecture uses this biological principle to leverage vision transformers to process only the most informative patches, allowing for a substantial reduction in memory usage that scales with the sequence length of selected patches. 
This approach aligns with visual neuroscience findings, suggesting that the human visual system optimizes information gathering through selective, spatially dynamic focus. 
Experimental evaluations on Imagenet-1k demonstrate that SMT achieves competitive top-1 accuracy while significantly reducing memory consumption and computational complexity, particularly when a limited number of patches is used. 
This work introduces a saccade-like selection mechanism into transformer-based vision models, offering an efficient alternative for image analysis and providing new insights into biologically motivated architectures for resource-constrained applications.

\keywords{sensorimotor systems  \and vision transformer \and saliency prediction.}
\end{abstract}
\section{Introduction}

Humans perceive their environment through a combination of various sensory inputs, with vision being one of the primary senses.
Unlike a camera that captures an entire image simultaneously, we perceive the world through rapid, jerky eye movements known as saccades.
These saccades shift our high-resolution central vision, the fovea, to points of interest in the environment.
This dynamic sampling process allows us to gather information and better understand the visual world efficiently.

Inspired by this biological strategy, we present a novel computer vision system that mimics human saccadic eye movements to understand images.
Our system leverages the power of Vision Transformer (ViT)~\cite{dosovitskiy2020vit} to extract meaningful information from images.
Firstly, we compute a saliency map, identifying the most visually interesting regions within the image.
These regions likely correspond to where humans would direct their gaze during saccades.
We then strategically extract image patches from the most salient locations and process them through a vision transformer architecture.
This approach allows the system to focus computational resources on the most informative parts of the image, similar to how the human visual system prioritizes information from the fovea.

The saliency operator is based on the concept of intrinsic dimensionality by Zetzsche~et~al.~\cite{Zetzsche1990}.
Intrinsic two-dimensional signals (i2D signals) are distinctive visual elements like corners, occlusions, and line endings. 
They are considered to be the most information-rich and least redundant parts of an image. 
They are less common in natural scenes than regions with constant luminance (i0D) or simple edges (i1D) but are crucial for processing complex visual information. Due to their low redundancy, i2D signals are prioritized in visual perception, influencing object recognition and triggering visual illusions such as the Kanisza triangle.
Furthermore, Krieger~et~al.~\cite{Krieger2000fixations} argue that i2D-points are good candidate locations for fixations by humans.

\subsection{Key Contributions}
The key contributions of this work are as follows:
\begin{enumerate}
    \item \textbf{Biologically Inspired Sensorimotor Framework in Vision Transformers:} This work draws from the human sensorimotor process, particularly saccadic eye movement, to introduce a biologically inspired perception-action model within vision transformers. 
    By simulating selective foveal fixations through patch-based attention, the model aligns computational processes with human visual strategies, improving focus on information-dense areas.
    \item \textbf{Saliency-Driven Patch Selection Mechanism:} To address the limitations of standard Vision Transformers (ViTs)~\cite{dosovitskiy2020vit} in handling large sequences, we introduce a saliency-based patch selection mechanism. 
    This feature allows the model to prioritize and process only the most informative image patches, reducing computational redundancy and enhancing model efficiency.
    \item \textbf{Improved Computational Efficiency and Memory Usage:} By implementing selective attention to salient patches, our approach reduces the computational and memory overhead of ViTs without sacrificing model performance, achieving an architecture that is both computationally efficient and biologically inspired.
\end{enumerate}

\section{Related Work}
While the concept of saccadic eye movements is gaining traction in computer vision, few published works directly implement them.
However, ongoing research utilizes related ideas like attention mechanisms and saliency maps to achieve similar goals.

\subsection{Vision Transformer}
Vaswani~et~al.~\cite{vaswani2017transformer} introduced the transformer architecture, which gained traction in the scientific community very fast.
The self-attention mechanism allows the model to focus on specific parts of the input based on their importance for the task.
While the original model was developed for text-to-text translation, the architecture was soon adopted in other domains.
Dosovitskiy~et~al.~\cite{dosovitskiy2020vit} applied the model to images, introducing the vision transformer (ViT).
There are many developments based on the ViT for which we would like to redirect to a review \cite{Khan2023vitreview} featuring the most important ones.

Our system is related to the class of masked autoencoders~\cite{He2022maskedautoencoders} and transformers where image patches are blacked out (see Zhang~\cite{zhang2022surveymaskedautoencoderselfsupervised} for a review on masked autoencoders).

Furthermore, there are models where a selection mechanism is used to prune input and improve computational efficiency. 
For example, various input sampling strategies have been introduced in \cite{chen2022}, \cite{ehteshami2022}, \cite{elsayed2019saccader}, \cite{khaki2024needspeedpruningtransformers}, \cite{kwan2022},  \cite{liu2024improved}. 
\cite{chen2022} and \cite{khaki2024needspeedpruningtransformers} list additional recent papers with sampling and pruning techniques for ViT.

\subsection{Models of Saccadic Eye Movements}
While there is a large amount of literature on human visual perception and saccadic eye movements (see \cite{Schuetz2011eyemovements} for a review), few models are trying to use the concept for computer vision.
Schill~et~al.~\cite{Schill01okusys} introduced a hybrid system that combined a knowledge-based reasoning system with low-level preprocessing by linear and nonlinear neural operators.
Farkya~et~al.~\cite{farkya2022saccade} explored mimicking saccades to improve efficiency in deep neural networks for classification and object detection.
It investigated using saccade-like mechanisms to mask irrelevant image patches and reduce computational demands.
Li~et~al.~\cite{li2022modelinghumaneyemovements} proposed an RNN-based model with separate CNNs for processing gaze and scene information, mimicking the human gaze shift during a maze exploration task.

\section{Model and Methods}

\subsection{Connection between Sensorimotor Systems and Transformers}
A sensorimotor system can be broadly conceptualized as a "perception-action" loop, in which sensory input guides motor actions and these actions subsequently modify future sensory input~\cite{GHAHRAMANI1997117}. 
Human saccadic eye movements represent a natural example of such a system. In this system, the eye fixates (perceives) on the point of interest in the environment, followed by a saccadic movement (action) that shifts fixation to a new area of interest. 
This cycle, which occurs in rapid succession, allows for efficient exploration and analysis of complex scenes.

Schill~et al.~\cite{Schill01okusys} formalizes this concept of sensorimotor feedback through a \textit{Sensorimotor Feature} (SMF), represented as a triplet 
\begin{equation}
    SMF := ( s_{t-1}, m_{t-1}, s_t)
\end{equation}
where \( s_{t-1} \) and \( s_t \) are sensory inputs at consecutive time points, and \( m_{t-1} \) is the motor action linking them. 
Here, the sensory input is spatially restricted, akin to the high-resolution focal area of the fovea in human vision. 
As the loop iterates over time, it generates a sequence of sensory inputs, building a progressively enriched understanding of the visual field.

This perception-action cycle closely resembles the transformer architecture, particularly in vision transformers (ViTs). 
In a ViT, an image is divided into discrete patches that form an ordered sequence akin to the sensory inputs in a sensorimotor system. 
The "action" component in transformers is embodied in the attention mechanism, where the attention matrix determines the "movement" from one patch to another. 
This patch-to-patch attention allows the transformer to dynamically shift focus across regions of an image, effectively simulating a virtual saccadic movement between patches. 

However, a standard ViT lacks a built-in mechanism for selectively focusing on salient patches. 
Each patch is processed in sequence, regardless of its information content, potentially leading to computational inefficiencies. 
In our work, we address this limitation by introducing a method to identify and prioritize salient patches. This enables a selective focus within the transformer framework that mimics the human visual system’s emphasis on high-information areas. 
This selection mechanism enhances the transformer’s efficiency and aligns its processing with biologically inspired strategies.

\subsection{System Overview}
\begin{figure}[!ht]
    \centering
    \includegraphics[width=0.7\linewidth]{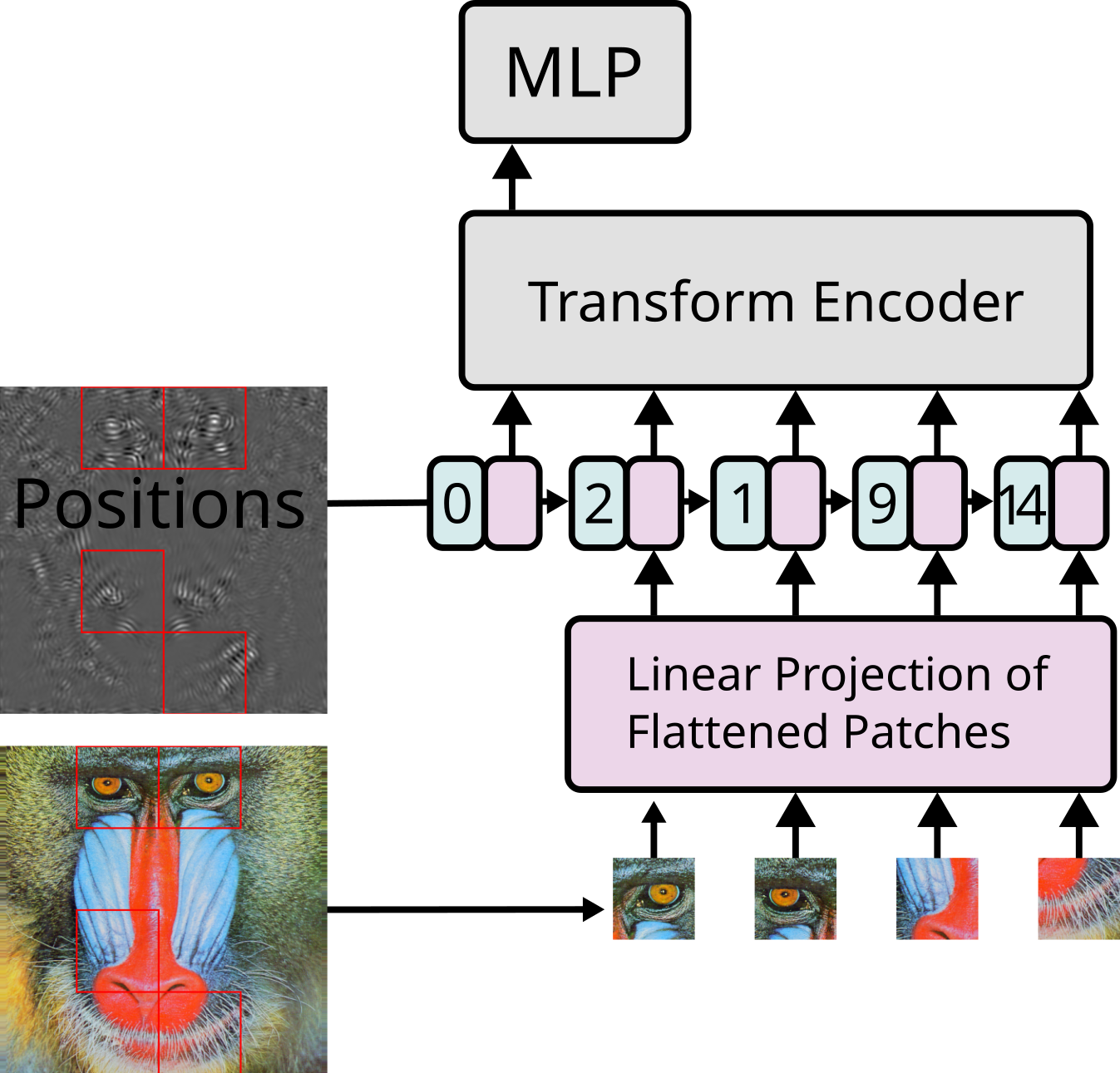}
    \caption{Overview of the proposed Sensorimotor Transformer}
    \label{fig:smt}
\end{figure}

Our system, the Sensorimotor Transformer (SMT), is based on the Vision Transformer (ViT) architecture~\cite{dosovitskiy2020vit}. 
Still, it introduces a novel approach by selecting and processing only the most salient patches instead of using all available patches in the input array. 
This selective focus enhances computational efficiency by prioritizing the most informative regions within the image. 
The absolute positions of these selected patches are encoded using a linear layer that captures their coordinates within the patch array, effectively retaining spatial information. 
These encoded positional embeddings, combined with the linear projections of the patch features, are then fed into the transformer blocks, where they are processed in the same manner as in the original Vision Transformer architecture. 
This approach maintains the advantages of the transformer’s self-attention mechanism while reducing the overall computational load by focusing on key visual information.
Figure~\ref{fig:smt} shows an overview of the proposed architecture.

Besides its biologically inspired design, the Sensorimotor Transformer (SMT) offers practical advantages in reducing computational complexity and memory usage for transformer-based models. 
Memory consumption and computational complexity in transformers predominantly scale with the length of the input sequence due to the quadratic complexity of the self-attention mechanism~\cite{Keles2023complexity}. 
By employing a limited number of image patches, the SMT effectively reduces the sequence length, correspondingly decreasing overall memory requirements and computational complexity. 
This selective patch approach efficiently processes essential visual information while avoiding the exhaustive memory demands typical of full patch-based models, making SMT an alternative for vision tasks where computational resources are constrained. 
Thus, the SMT leverages biological principles to improve focus on salient features and optimize memory utilization in deep learning applications.

\subsection{Prediction of Salient Regions}
The basic idea behind predicting salient regions is rooted in the observation that intrinsically two-dimensional (i2D) points are likely candidates for fixation locations~\cite{Krieger2000fixations}. 
Such i2D points can be effectively identified through the concept of Gaussian curvature, as proposed in prior studies by Zetzsche~et al.~\cite{Zetzsche1990}, \cite{Zetzsche1990a}. 
Gaussian curvature provides a measure of surface shape that distinguishes points with high informational relevance. 
Multiple approaches to computing curvature can be implemented in both spatial and frequency domains, as outlined in~\cite{ZETZSCHE2001intrinsic_dimensionality}. 
For this work, we utilized the Iso-Curvature filter developed in \cite{ZETZSCHE2001intrinsic_dimensionality}, which has shown robustness in identifying key i2D points and requires few filter operations. 
In the following subsections, we will present the complete filter chain employed for predicting saliency, detailing each process step.

\subsubsection{Image contrast}
The first step in the model is to abstract from the absolute luminance of the input signal, achieving invariance to luminance variations, a common objective in computational models of the human visual system. 
For this purpose, we use the Ratio-of-Gaussians (RoG) operator to compute contrast, as outlined in Eq.~\eqref{eqn:rog}. 

\begin{equation}
    g_{i+1}^{RoG}(x,y) = \frac{l(x,y) * \frac{1}{2 \pi \sigma_{i}^2} \exp \bigg(-\left(\frac{x^2 + y^2}{2\sigma_{i}^2}\right)\bigg) }{ \Bigg( l(x,y) * \frac{1}{2 \pi \sigma_{i+1}^2} \exp \bigg(-\left(\frac{x^2 + y^2}{2\sigma_{i+1}^2}\right)\bigg) \Bigg) + \tau} \label{eqn:rog}
\end{equation}

\begin{figure}[!ht]
    \centering
    \begin{subfigure}{0.3\linewidth}
    \centering
    \includegraphics[width=1\linewidth]{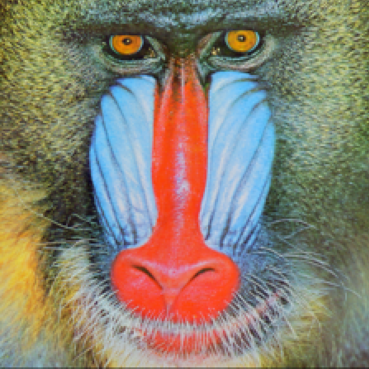}
    \caption{Input image.}
    \label{fig:rog_input}
  \end{subfigure}
  \hfill
  \begin{subfigure}{0.3\linewidth}
    \centering
    \includegraphics[width=1\linewidth]{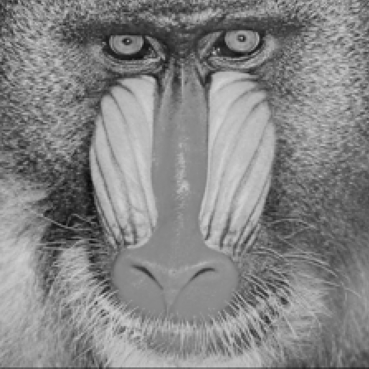}
    \caption{Luminance.}
    \label{fig:rog_lum}
  \end{subfigure}
  \hfill
        \begin{subfigure}{0.3\linewidth}
        \centering
    \includegraphics[width=1\linewidth]{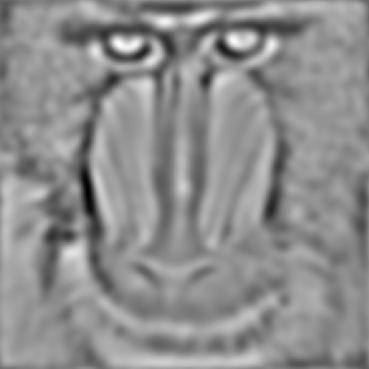}
    \caption{Local contrast.}
    \label{fig:rog_contrast}
  \end{subfigure}        
        \caption{Ratio of Gaussians contrast operator.}
        \label{fig:rog}
    \end{figure}

Unlike the traditional Difference-of-Gaussians (DoG) operator~\cite{Enroth1966}, \cite{Enroth1983}, \cite{Rodieck1965}, which computes contrast by subtracting two Gaussian-blurred versions of the signal, the RoG operator divides the outputs of the two Gaussians. 
This division-based approach enables RoG to operate directly on the luminance signal without requiring the input to be transformed into logarithmic space. 
Thus, the luminance signal's natural characteristics are preserved while providing robust contrast information. 
This method enhances the model's ability to mimic the luminance-invariance observed in human vision, setting a solid foundation for subsequent processing steps.
Figure~\ref{fig:rog} illustrates the contrast computation.

\subsubsection{Curvature Operator}
Generally, intrinsically two-dimensional (i2D) signals require an AND-like combination of filter responses for detection, as linear filters alone are insufficient to capture this class of signals~\cite{Zetzsche1990}. 
Detecting i2D signals can be achieved through the concept of Gaussian curvature, classified by calculating the determinant of the Hessian matrix

\begin{equation}
    D = l_{xx} \ l_{yy} - l_{xy}^2
\end{equation}

where \( l_{xx} \), \( l_{yy} \), and \( l_{xy} \) represent derivatives in the \( xx \), \( yy \), and \( xy \) directions, respectively~\cite{Zetzsche1990a}. 
Here, \( l_{xx} \) and \( l_{yy} \) correspond to orientation-selective filters with distinct preferred orientations combined in an AND-like manner to capture i2D signals. 
However, due to the partial overlap in their orientation sensitivity, these filters may also respond to intrinsically one-dimensional (i1D) signals that align within the overlapping orientation range. 
The term \( l_{xy} \) addresses this overlap by acting as a rotated filter that selectively targets the overlapping region, allowing compensation for i1D signals~\cite{Zetzsche1990a}.
By subtracting \( l_{xy}^2 \) from the AND-like combination, the model enhances its selectivity for genuine i2D signals, minimizing interference from i1D responses.

\paragraph{Filter Design for i2D Selectivity}

As described above, i2D selectivity can be achieved through an AND-like combination of filter responses with differing orientation tunings, effectively capturing intrinsically two-dimensional (i2D) features in visual data. 
While directional filters can be used, isotropic filters, if they have different orientation tuning characteristics, are equally effective for this purpose. 
Here, we implement the isotropic curvature operator introduced in Zetzsche~\cite{Zetzsche1990a}.

In this section, we provide a detailed description of our filter implementation. 
The polar-separable filters enable efficient decomposition into radial and angular components within a polar coordinate system. 
In this framework, polar coordinates \( (r, \phi) \) are defined as:

\begin{equation}
    r = \sqrt{x^2 + y^2}, \quad \phi = \arctan(y, x)
\end{equation}
    
where \( r \) is the radial distance from the origin, and \( \phi \) represents the angle in polar coordinates.

We need a radial and two orientation-selective angular components to compute the curvature. 

The radial component, \( g(r) \), is a Gaussian function defined on the radius \( r \):

\begin{equation}
    g(r) = (2 \pi r)^2 \exp \left( -\frac{\pi r^2}{\sigma^2} \right)
\end{equation}

This Gaussian function provides smooth isotropic filtering across the radial dimension. 
The orientation filter function, on the other hand, is defined in terms of sinusoidal functions applied to the polar angle \( \phi \), using both cosine and sine components to capture oscillatory angular patterns. 
Here, \( n \) represents the number of oscillations, and \( i \) denotes the imaginary unit:

\begin{equation}
    c_n(\phi) = i^n \cos(n \, \phi)
\end{equation}
\begin{equation}
    s_n(\phi) = i^n \sin(n \, \phi)
\end{equation}
    
These radial and orientation components are applied to the Fourier-transformed image, with the zero-frequency shifted to the center for easy filtering. 
The Ratio-of-Gaussians (RoG) filter first processes the image to compute local contrast and approximate luminance invariance.

The resulting filtered components are as follows:
\begin{itemize}
    \item     Cosine component
    \begin{equation}
    C_n(r, \phi) = G^{RoG}(r, \phi) \ c_n(\phi) \ g(r)    
    \end{equation}

    \item Sine component
    \begin{equation}
        S_n(r, \phi) = G^{RoG}(r, \phi) \ s_n(\phi) \ g(r)
    \end{equation}

    \item Laplace component
    \begin{equation}
        Laplace(r, \phi) = G^{RoG}(r, \phi) \ g(r)
    \end{equation}
\end{itemize}

\begin{figure*}
    \centering
    \begin{subfigure}{0.3\linewidth}
    \includegraphics[width=1\linewidth]{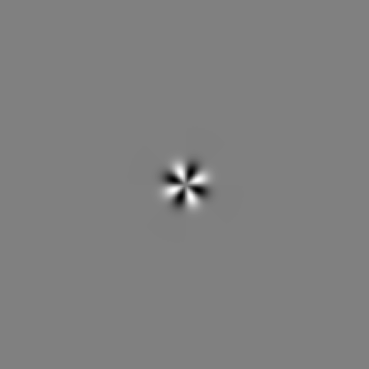}
    \caption{$S_n$}
    \label{fig:s_n}
  \end{subfigure}
  \hfill
      \begin{subfigure}{0.3\linewidth}
    \includegraphics[width=1\linewidth]{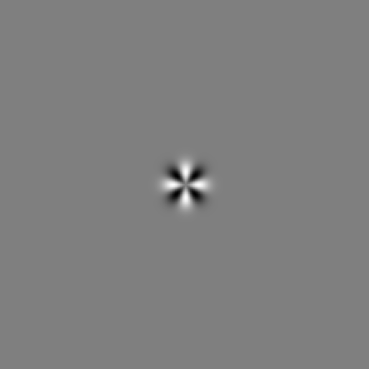}
    \caption{$C_n$}
    \label{fig:c_n}
  \end{subfigure}
  \hfill
      \begin{subfigure}{0.3\linewidth}
    \includegraphics[width=1\linewidth]{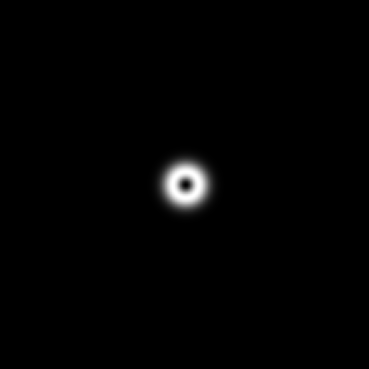}
    \caption{$Laplace$}
    \label{fig:laplace}
  \end{subfigure}

\caption{Polar separable filter functions. \protect\subref{fig:s_n} and \protect\subref{fig:c_n} show the angular filter functions and \protect\subref{fig:laplace} the radial one.}
\label{fig:filter_functions}
    
\end{figure*}

These filtered outputs allow us to detect i2D features with radial and angular specificity, enabling selective detection of salient regions in an image. 
This approach leverages isotropic and orientation-selective filtering to approximate human visual sensitivity to intrinsic two-dimensional structures, facilitating effective i2D signal capture within the proposed model.
Figure~\ref{fig:filter_functions} visualizes the filter functions in the Fourier domain for $n=2$.

\paragraph{Curvature Computation}

The eccentricity, denoted as \( Ecc_n \), represents the norm of orientation-selective filters applied across \( n \) orientations. 
This measure captures the combined response strength of the cosine and sine components at each polar coordinate \((r, \phi)\), and is defined as:

\begin{equation}
    Ecc_n(r, \phi) = \sqrt{C_n(r, \phi)^2 + S_n(r, \phi)^2}
\end{equation}
    
The curvature at each point \((r, \phi)\) is then calculated by comparing the squared response of the Laplace component with the squared eccentricity. 
This formulation provides a measure of curvature based on the difference in response strength between isotropic (Laplace) and directional (eccentricity) components:

\begin{equation}
    D(r, \phi) = Laplace(r, \phi)^2 - Ecc_n(r, \phi)^2
\end{equation}
    
\begin{figure*}
    \centering
    \begin{subfigure}{0.3\linewidth}
    \includegraphics[width=1\linewidth]{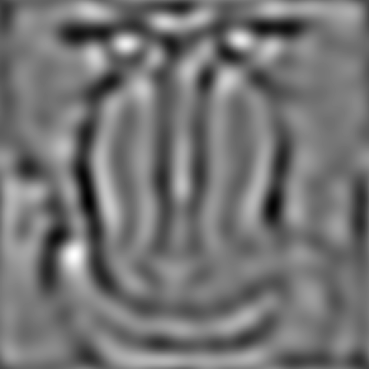}
    \caption{$Laplace$}
    \label{fig:laplace_response}
  \end{subfigure}
  \hfill
\begin{subfigure}{0.3\linewidth}
    \includegraphics[width=1\linewidth]{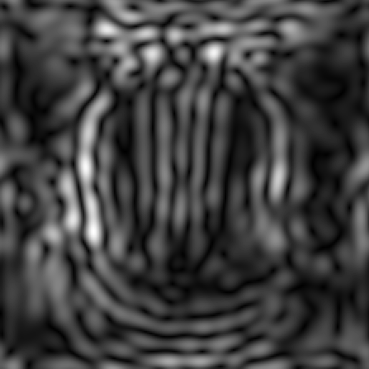}
    \caption{$Ecc_n$}
    \label{fig:ecc_response}
  \end{subfigure}
  \hfill
    \begin{subfigure}{0.3\linewidth}
    \includegraphics[width=1\linewidth]{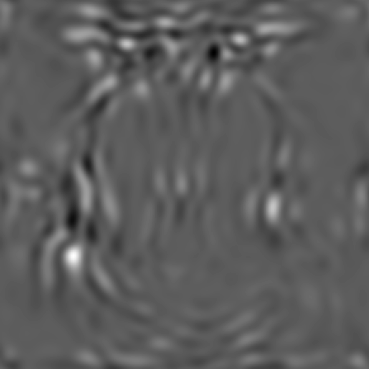}
    \caption{$D$}
    \label{fig:D_response}
  \end{subfigure}
  \hfill

    \caption{Filter responses of the curvature components \protect\subref{fig:laplace_response} and \protect\subref{fig:ecc_response} and the final curvature \protect\subref{fig:D_response}.}
\label{fig:curvature}
\end{figure*}

For our application, it is advantageous to take the absolute value of the curvature,
\begin{equation}
    D_{abs}(r, \phi) = | D(r, \phi)|
\end{equation}
as the direction of curvature (positive or negative) is not of primary importance. 
Instead, the focus is on the magnitude of curvature, which reflects the strength of intrinsic two-dimensional features within the image. 
This approach emphasizes the overall curvature intensity, making it suitable for applications prioritizing salient structural information.
Figure~\ref{fig:curvature} shows exemplary filter responses for curvature computation.

\subsubsection{Patch selection}

To effectively identify and prioritize the most informative regions in the curvature image, we employ a patch-based selection process inspired by the standard Vision Transformer architecture~\cite{dosovitskiy2020vit}. 
First, the curvature image is divided into non-overlapping square patches of size \( p \times p \). 
Each patch represents a localized region of the curvature image, allowing for regional analysis of the curvature response.

Once the image is divided into patches, we compute the sum of curvature responses within each patch. 
This summed response measures each patch's significance and reflects the overall strength of curvature in that region. 
After calculating these summed responses for all patches, they are sorted in descending order to prioritize areas with the highest curvature, corresponding to regions with potentially high visual salience.

We select the top \( m \) patches from the sorted list, which exhibit the highest summed curvature values. 
These \( m \) selected patches represent the regions most likely to contain key structural and feature-rich content, making them ideal for subsequent processing within the Vision Transformer pipeline. 
This selection process improves computational efficiency by focusing on the most salient parts of the image, aligning with our goal of reducing input redundancy while preserving critical information.

\subsection{Model Implementation, Parameters, and Training}

The model implementation was carried out in PyTorch 2.1, using the excellent \texttt{timm }framework\footnote{https://github.com/huggingface/pytorch-image-models}, known for its high-quality support of a wide range of vision models. 
The filters are implemented in PyTorch as well using PyTorch's FFT.

In our experiments, we used the base configuration of the Vision Transformer (ViT) model, adopting a patch size \( p = 16 \) for input partitioning, as outlined in Table~\ref{tab:model_parameters}. The number of selected patches, \( s \), was varied between 49, 98, 147, and 196, corresponding to 25\%, 50\%, 75\%, and 100\% of the total patches, respectively, for an input image size of 224x224 pixels.

\begin{table}[h]
    \centering
    \caption{Model parameters corresponding to "base" version of ViT~\cite{dosovitskiy2020vit}.}
    \label{tab:model_parameters}
    \begin{tabular}{@{}lc@{}}
    \toprule
    Parameter & Value\\
    \midrule
        Input size          & 224 \\
         Dense dim          &   768  \\
         Number of heads    &   12   \\
         Depth              &   12  \\
         MLP dimension      &   3072    \\
         Patch size         &   16  \\
         \bottomrule
    \end{tabular}    
    
\end{table}

The training was conducted on the ImageNet1k dataset~\cite{imagenet_cvpr09}, with inputs normalized following the "Inception" normalization protocol~\cite{Carreira17inception}. 
Specific training parameters are presented in Table~\ref{tab:training_parameters}.

\begin{table}[]
    \centering
    \caption{Training parameters.}
    \label{tab:training_parameters}
    \begin{tabular}{@{}lc@{}}
    \toprule
    Parameters & \\
    \midrule
         Number of epochs   & 300 \\
         Learning rate      &   0.003 \\
         Optimizer          &   AdamW \\
         Warmup Steps       &   20\\
         Scheduler          &   Cosine \\
         Dropout            &   0.1 \\
         Weight Decay       &   0.3 \\
         Gradient Accumulation  & 4096  \\
         Augmentation       & none  \\
         Clip Gradients     & true  \\
         \bottomrule
    \end{tabular}

\end{table}

\section{Results}
This section presents the initial results from our Sensorimotor Transformer (SMT) model. 
The primary objective was establishing stable training performance rather than achieving state-of-the-art (SOTA) accuracy, so the training parameters were not optimized for maximum accuracy.

\begin{figure}
    \centering
    \includegraphics[width=0.7\linewidth]{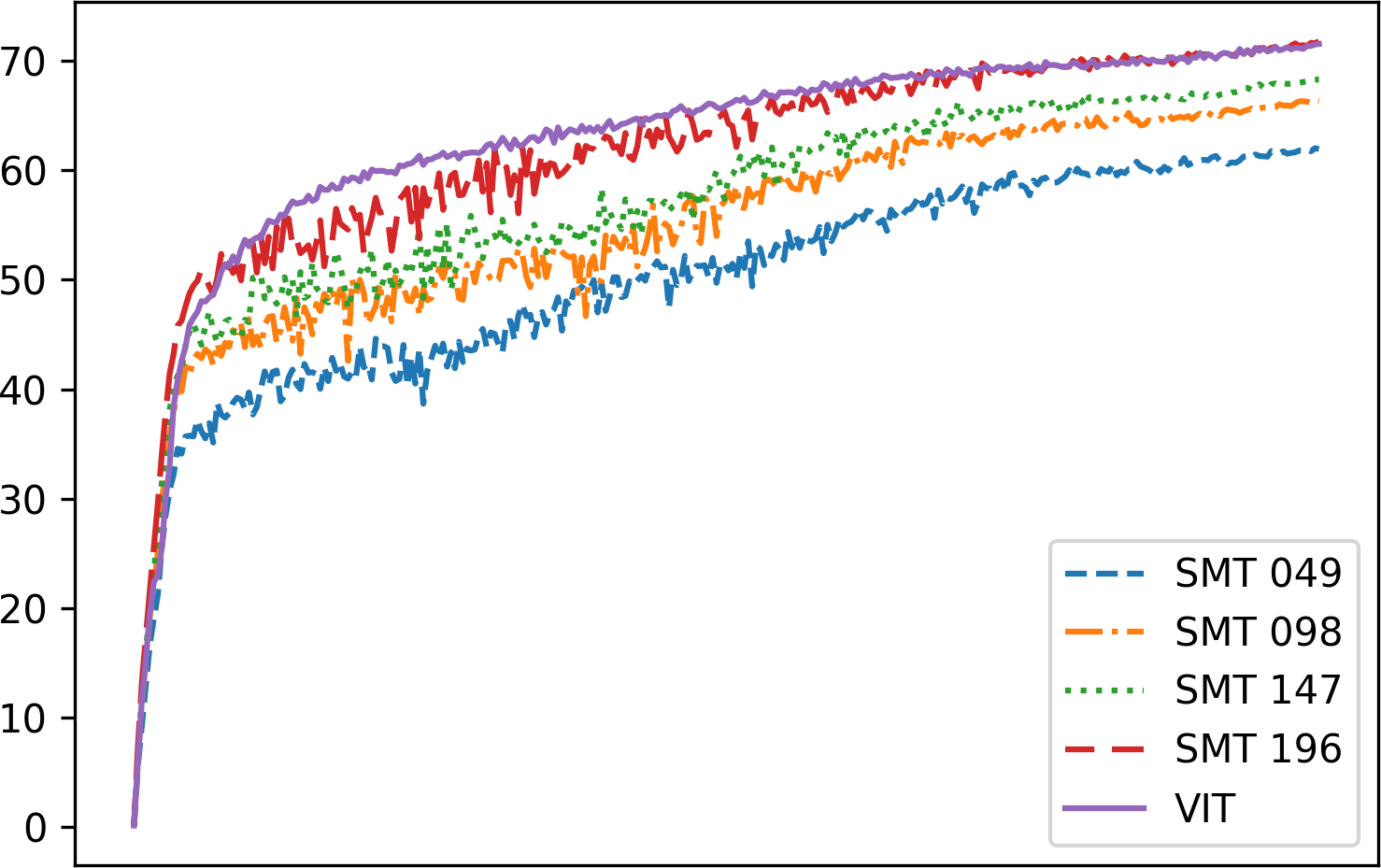}    
    \caption{Top 1 accuracy.}
\label{fig:top1_accuracy}
\end{figure}

\paragraph{Model Accuracy}
We evaluated the SMT's top-1 accuracy in the ImageNet-1k data set, comparing its performance against the Vision Transformer (ViT) baseline. 
Performance was measured across different SMT configurations by varying the number of patches used (49, 98, 147, and 196). 
Table~\ref{tab:top1_acc_results} provides detailed results, while Figure~\ref{fig:top1_accuracy} visualizes the comparative top-1 accuracy across models.

\begin{table*}[h]
    \centering    
    \caption{Top-1 Accuracy on ImageNet-1k: Comparison of SMT and ViT under different training configurations.}
    \label{tab:top1_acc_results}
    \begin{tabular}{@{}lc@{}}
    \toprule
    Model & Accuracy (\%)\\
    \midrule
         ViT         &  71.5  \\
         \hline
         SMT$_{49}$   & 61.9  \\
         SMT$_{98}$   & 66.3     \\
         SMT$_{147}$  & 68.3  \\
         SMT$_{196}$  & 71.8  \\
         \bottomrule
    \end{tabular}       
    
\end{table*}

In general, the SMT models achieved slightly lower accuracy than the ViT, with SMT$_{196}$ (using 100\% of patches) approaching the baseline accuracy of the ViT. 
The performance showed a progressive decline with fewer patches, with SMT$_{49}$ (25\% of patches) showing the most significant accuracy drop, highlighting the dependence on patch density for effective feature representation.

\paragraph{Model Memory Consumption}
We also evaluated the memory consumption of the SMT configurations during training using a batch size of 256, as shown in Table~\ref{tab:memory_results}. 
Memory usage scaled predictably with the number of patches, though a fixed overhead resulted in slightly less-than-perfect proportional scaling. 
The SMT ${49}$ configuration used only 31.0\% of the memory of SMT$_{196}$, making it a highly efficient alternative regarding GPU memory requirements.

\begin{table*}[h]
    \centering
    \caption{Memory consumption on GPU during training with batch size 256 for different SMT configurations.}
    \label{tab:memory_results}
    \begin{tabular}{@{}lcc@{}}
    \toprule
    Model & Memory Usage (MiB) & Percentage of SMT$_{196}$ Usage \\
    \midrule
         SMT$_{49}$   & 9,896     & 31.0\% \\
         SMT$_{98}$   & 17,396    & 54.5\%    \\
         SMT$_{147}$  & 24,898    & 78.0\%      \\
         SMT$_{196}$  & 31,932    & 100.0\%    \\
         \bottomrule
    \end{tabular}       
    
\end{table*}

These results indicate that the SMT model's memory consumption scales roughly with patch count, providing a flexible trade-off between computational efficiency and model performance based on available resources and specific task requirements.

\section{Conclusion}

In this work, we introduced the Sensorimotor Transformer (SMT), a biologically inspired computer vision model that mimics the human strategy of saccadic eye movements by selectively attending to the most salient regions within an image. 
Building on principles from human vision, such as prioritizing high-information i2D signals and dynamic sampling achieved through saccades, the SMT efficiently reduces computational demands by limiting the number of processed image patches. 
The SMT maximizes accuracy by focusing on informative regions with a transformer-based architecture while maintaining lower memory usage than traditional models that process full-image patches. 
Experimental results demonstrate that this selective patch approach achieves competitive performance on Imagenet-1k, with a notable reduction in memory requirements that scales with the length of the patch sequence. 
This biologically inspired approach thus highlights the potential for saccade-like patch selection to improve efficiency and performance in deep learning vision models, setting a promising direction for further research on adaptive, resource-efficient computer vision systems.

\subsubsection*{Acknowledgements} 
The research reported in this paper has been supported by the German Research Foundation DFG, as part of Collaborative Research Center (Sonderforschungsbereich) 1320 Project-ID 329551904 “EASE - Everyday Activity Science and Engineering”, University of Bremen (http://www.ease-crc.org/). The research was conducted in subproject H03 "Discriminative and Generative Human Activity Models for Cognitive Architectures".
%
%
%
\bibliographystyle{splncs04}
\bibliography{references}
\end{document}